\documentclass[10pt,twocolumn,letterpaper]{article}

\usepackage{cvpr}
\usepackage{times}
\usepackage{epsfig}
\usepackage{graphicx}
\usepackage{amsmath}
\usepackage{amssymb}
\usepackage{color}
\usepackage{tabularx}
\usepackage{multirow}
\usepackage{booktabs}
\usepackage{enumitem}
\usepackage[export]{adjustbox}

\usepackage{mwe} 
\usepackage{subcaption}

\usepackage[pagebackref=true,breaklinks=true,letterpaper=true,colorlinks,bookmarks=false]{hyperref}

\cvprfinalcopy 

\def\cvprPaperID{2052} 

\ifcvprfinal\fi
\begin{document}

\title{Aerial Spectral Super-Resolution using Conditional Adversarial Networks}

\author{Aneesh Rangnekar \qquad Nilay Mokashi \qquad Emmett Ientilucci \\ Christopher Kanan \qquad Matthew Hoffman \\
Rochester Institute of Technology \\
{\tt\small \{aneesh.rangnekar, nilaymokashi\}@mail.rit.edu, emmett@cis.rit.edu, \{kanan, mjhsma\}@rit.edu}
}

\maketitle

\begin{abstract}
Inferring spectral signatures from ground based natural images has acquired a lot of interest in applied deep learning. In contrast to the spectra of ground based images, aerial spectral images have low spatial resolution and suffer from higher noise interference. In this paper, we train a conditional adversarial network to learn an inverse mapping from a trichromatic space to 31 spectral bands within 400 to 700 nm. The network is trained on AeroCampus, a first of its kind aerial hyperspectral dataset. AeroCampus consists of high spatial resolution color images and low spatial resolution hyperspectral images (HSI). Color images synthesized from 31 spectral bands are used to train our network. With a baseline root mean square error of $2.48$ on the synthesized RGB test data, we show that it is possible to generate spectral signatures in aerial imagery. 
\end{abstract}

\begin{figure}[ht]
\begin{center}
\includegraphics[width=\linewidth, height = 10cm]{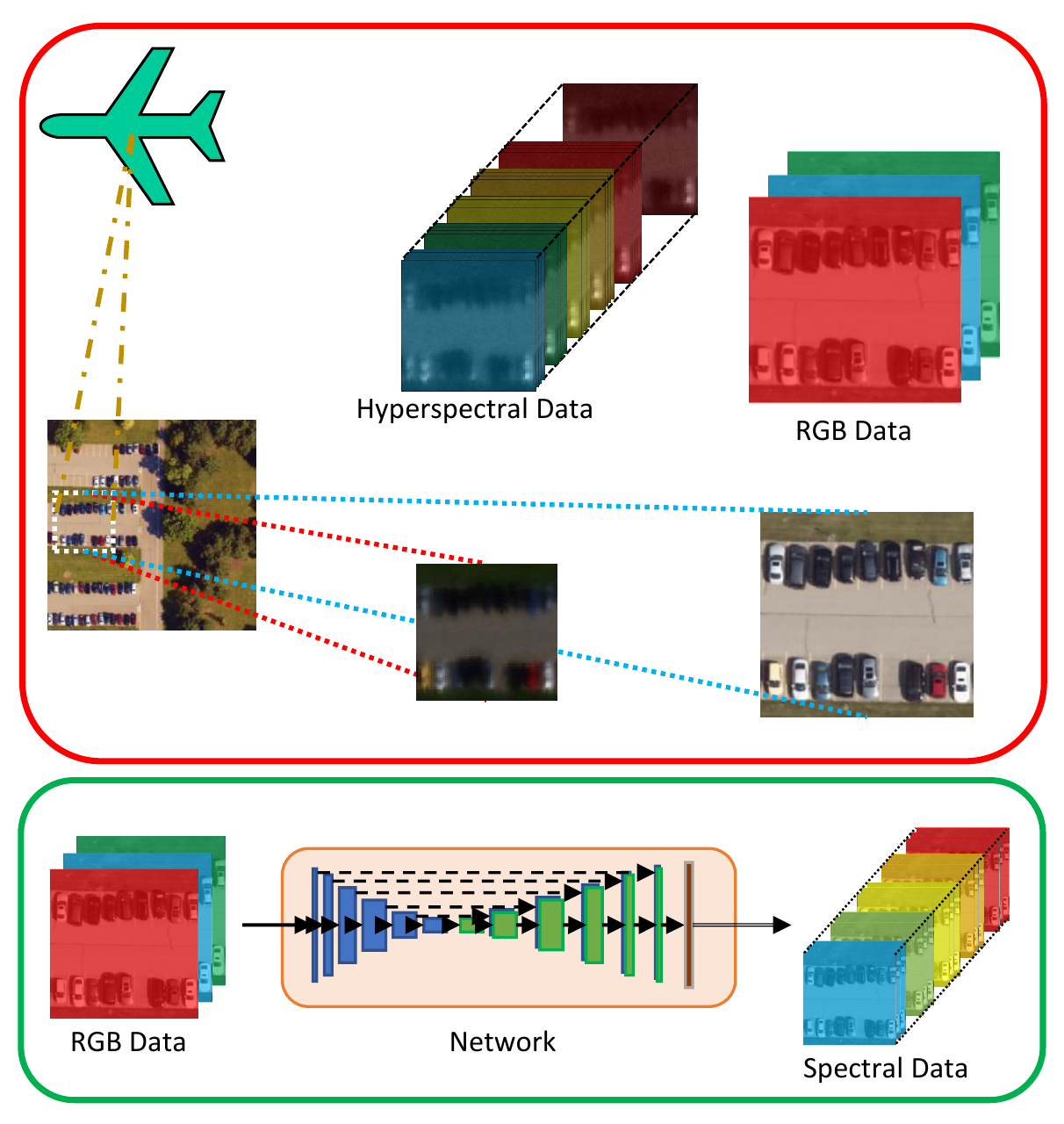}
\end{center}
   \caption{\textbf{Top.} RGB cameras provide high spatial resolution while hyperspectral cameras have low-spatial resolution that makes tasks significantly difficult. \textbf{Bottom.} In our approach, we infer a many channeled spectral image from an RGB image and to do this, we use conditional adversarial generative networks. The outcome is an image with \emph{both} high spatial resolution and high spectral resolution.}
\label{fig:flow}
\end{figure}

\section{Introduction}

Almost all consumer cameras available today function by converting the light spectrum to match the trichromaticity of the human eyes (as Red, Green and Blue channels). This is effective for presenting information to humans, but it ignores much of the visible spectrum. Hyperspectral images (HSI) and multispectral images (MSI),  on the other hand,  capture additional frequencies of the spectrum and often measure spectra with greater fidelity. This additional information can be used for many applications, including  precision agriculture~\cite{mulla2013twenty}, food quality analysis~\cite{sun2016computer} and aerial object tracking~\cite{uzkent2017aerial}. 

Typically, MSI have 4 - 10 channels spread over a large bandpass, and HSI have 30 - 600 channels with finer spectral resolution. MSI and HSI data can enable discrimination tasks where RGB will fail due to the increased spectral resolution. However, MSI and HSI data have drawbacks: (1) MSI and HSI cameras are very expensive, and (2) HSI and MSI have a significantly lower spatial and temporal resolution than RGB cameras (Fig.~\ref{fig:flow}). As a result, the use of spectral imagery has been limited to domains where these drawbacks are mitigated. Given the high hardware costs of flying an HSI  sensor, we explore the possibility of learning RGB to HSI mappings in low resolution spectral imagery and then applying those mappings to high resolution spatial RGB imagery to obtain images with \emph{both} high spatial and high spectral resolution. 

Spectral super-resolution SSR algorithms attempt to infer additional spectral bands in the 400 nm -- 700 nm range from an RGB and low resolution HSI images at an interval of 10 nm.  Recently, SSR algorithms using deep learning~\cite{galliani2017learned,alvarez2017adversarial,xiong2017hscnn} have been proposed that attempt to solve this problem in natural images. These methods bypass the need for a low resolution HSI input by learning RGB to Spectral mappings from a large sample of natural images~\cite{arad2016sparse}.

Recently, generative adversarial networks (GAN)~\cite{goodfellow2014generative} and its variants have shown tremendous success in being able to generate realistic looking images by learning a generative model of the data. Conditional GANs are similar to conventional GANs, except that they learn the output distribution as a function of noise and the input, thus making them suitable for text-to-image~\cite{zhang2016stackgan} and image-to-image~\cite{isola2016image} translation purposes. 

\textbf{This paper makes three major contributions}:
\begin{itemize}[noitemsep, nolistsep]
\item We show that conditional GANs can learn the target distribution for 31 spectral bands from low spatial resolution RGB images.  

\item We describe a new aerial spectral dataset called AeroCampus that contains a wide variety of objects, including, but not limited to, cars, roads, trees, and buildings.

\item We demonstrate that our conditional GAN achieves an effective root mean square error (RMSE) on AeroCampus of less than 3.0. We then use our model on RGB images with high spatial resolution to obtain images with both high spatial and high spectral resolution. 

\end{itemize}

\section{Related work}
\label{rw}

SSR is closely related to hyperspectral super-resolution~\cite{lanaras2015hyperspectral,akhtar2016hierarchical,dian2017hyperspectral}. Hyperspectral super-resolution involves inferring a high resolution HSI from two inputs: a low resolution HSI and a high resolution image (typically RGB). SSR is a harder task because it does not have access to the low resolution HSI, which can be expensive to obtain.

Nguyen \etal \cite{nguyen2014training} used a radial basis function (RBF) that leverages RGB white-balancing to recover the mapping from color to spectral reflectance values. They have two key assumptions that make their approach too restrictive: (1) They assume the color matching function of the camera is known beforehand and, (2) that the scene has been illuminated by an uniform illumination. Their method includes stages for recovering two things - the object reflectance and, the scene illumination and is very dependent on the assumptions for training the RBF network. Arad and Ben-Shahar \cite{arad2016sparse} proposed learning a sparse dictionary of hyperspectral signature priors and their corresponding RGB projections. They then used a many-to-one mapping technique for estimating hyperspectral signatures in the test image, while using all other images in the dataset for learning the dictionary. This approach yielded better results in \textit{domain-specific} subsets than the complete set uniformly since the dictionary has access to a lot similar naturally-occurring pixel instances in the training data and can be optimized for the target subset. Similar to Arad and Ben-Shahar, Aeschbacher \etal \cite{aeschbacher2017defense} adapted the A+ method \cite{timofte2014a+} to the spectral reconstruction domain to achieve significantly better results without the need for online learning of the RBG-HSI dictionary (Arad and Ben-Shahar's approach was inspired by the works of Zeyde \etal \cite{zeyde2010single}). However, these approaches tackle the mapping problem on a pixel level and fail to take advantage of area around the pixel that would possibly yield better information for predicting signatures, for example - if a particular color `blue' to be spectrally up-sampled, does it belong to the blue car or the sky? The above approaches fail to use this spatial information.

A number of papers that use applied deep learning for SSR have been published this year. Galliani \etal \cite{galliani2017learned} proposed the use of the \textit{Tiramisu} architecture \cite{jegou2017one}, a fully convolutional version of DenseNet \cite{huang2016densely}. They modified the network to a regression based problem by replacing Softmax-Cross Entropy loss for class segmentation with the Euclidean loss and established the first state-of-the-art results in the field. Xiong \etal proposed to use spectral interpolation techniques to first up-sample the RGB image in the channel space to a desired spectral resolution and then use CNNs to enhance the up-sampled spectral image. Similar to our work, Alvarez-Gila \etal \cite{alvarez2017adversarial} recently used a pix2pix \cite{isola2016image} image-to-image translation framework for SSR using GANs on natural images. A key point in applied deep learning methods being: unlike dictionary based algorithms which require information about the camera's color matching functions, these methods do not rely on this information. 

\section{AeroCampus RGB and HSI Data Sets} \label{datsetdescription}

\begin{figure}[t]
\begin{center}
\includegraphics[width=\linewidth]{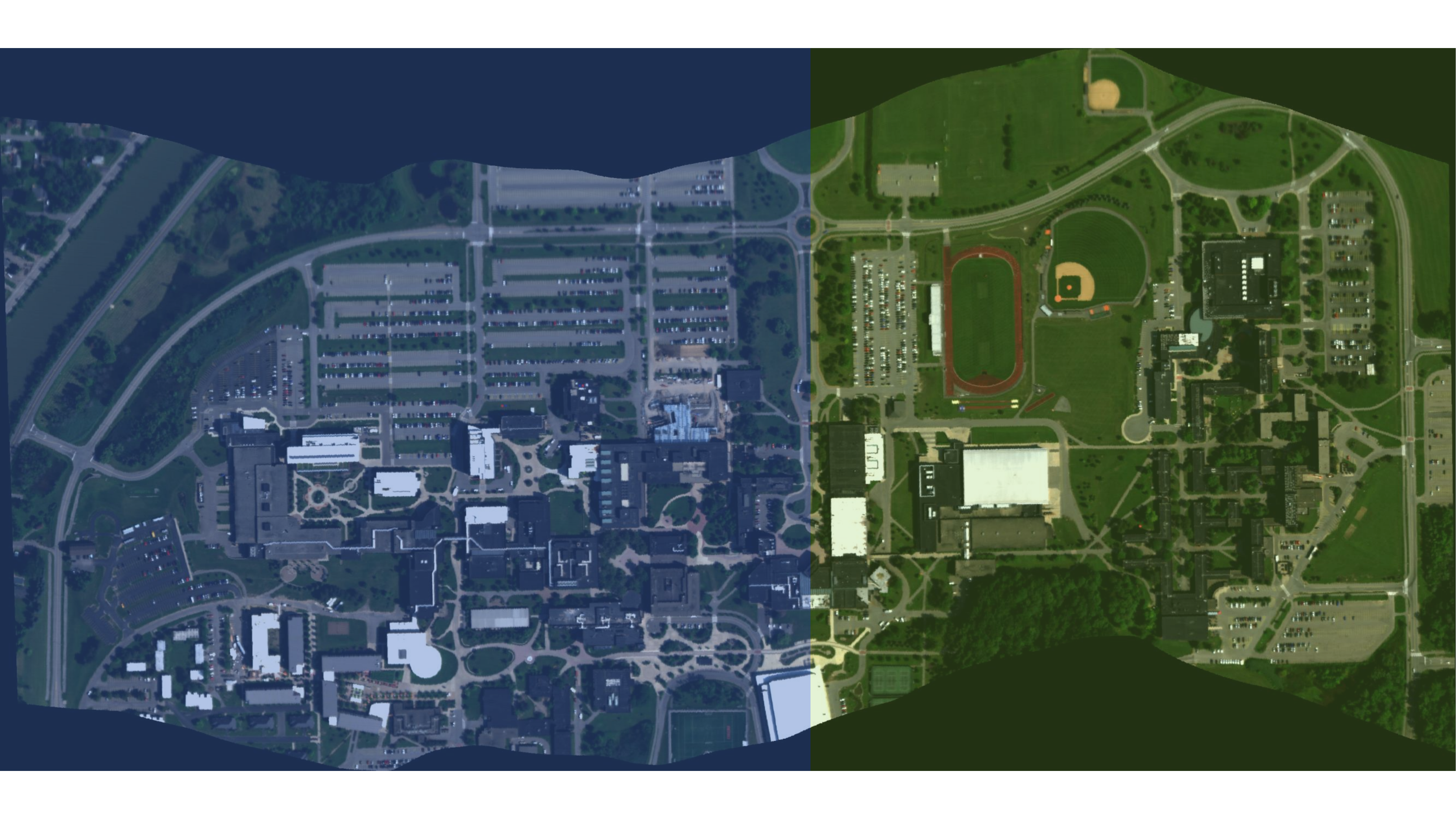}
\end{center}
   \caption{Geometrically corrected AeroCampus aerial flight line over Rochester Institute of Technology's university campus.  The image is segmented such that the right portion, shaded green, is used for testing while the left portion, shaded blue, is used for training.}
\label{fig:aeroRITone}
\end{figure}

The AeroCampus data set (see Fig.~\ref{fig:aeroRITone}) was generated by flying two types of camera systems over Rochester Institute of Technology's university campus on August 8th, 2017.  The systems were flown simultaneously in a Cessna aircraft.  The first camera system consisted of an 80 megapixel (MP), RGB, framing-type silicon sensor while the second system consisted a visible/near infrared (VNIR) hyperspectral Headwall Photonics Micro Hyperspec E-Series CMOS sensor. The entire data collection took place over the span of a couple hours where the sky was completely free of cloud cover, with the exception of the last couple flight lines at the end of the day.

The wavelength range for the 80 MP sensor was 400 to 700nm with typical band centers around 450, 550, and 650nm and full-width-half-max (FWHM) values ranging from 60-90nm.  The hyperspectral sensor provided spectral data in the range of 397 to 1003nm, divided into 372 spectral bands. The ground sample distance (GSD) is completely dependent on flying altitude. The aircraft was flown over the campus at altitude of approximately 5,000 feet, yielding an effective GSD for the RGB data of about 5cm and 40cm for the hyperspectral imagery. 

Both data sets were ortho-rectified based on survey grade GPS.  That is, camera distortion was removed along with uniform scaling and re-sampling using a nearest neighbor approach so as to preserve radiometric fidelity.  The RGB data was ortho-rectified onto the Shuttle Radar Topography Mission (SRTM) v4.1 Digital Elevation Model (DEM) while the HSI was rectified onto a flat plane at the average terrain height of the flight line (i.e., a low resolution DEM). Both data sets were calibrated to spectral radiance in units of $W m^{-2} sr^{-1} \mu m^{-1}$. To preserve the integrity of the training and testing data, we only use one of the six flight lines collected to record our results. There was significant overlap between the other flight lines and hence, the one with the largest spatial extent was chosen to obtain a considerable split in the dataset (Fig. \ref{fig:aeroRITone}). 


\begin{figure}[htb!]
\begin{center}
\begin{subfigure}[t]{0.5\linewidth}
        \centering
        \includegraphics[height=1.2in]{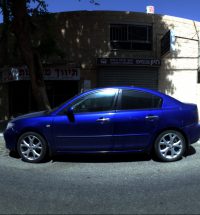}
        \caption{Blue Car (ICVL \cite{arad2016sparse})}
    \end{subfigure}%
    ~ 
    \begin{subfigure}[t]{0.5\linewidth}
        \centering
        \includegraphics[height=1.2in]{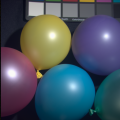}
        \caption{Balloons (CAVE \cite{yasuma2010generalized})}
    \end{subfigure}
\end{center}
   \caption{Left: Unique objects that occur only once in the spectral datasets and hence make it difficult to infer their signatures.}
\label{fig:guessornot?}
\end{figure}

\textbf{Comparison to other datasets.} To the best of our knowledge, AeroCampus is the first of its kind as an aerial spectral dataset. The closest contender would be the Kaggle DSTL Satellite Imagery Dataset with a 8 band multispectral channel between 400 nm to 1040nm. Not having an uniform pre-defined split also causes a problem when it comes to validating the current state of the art methods over newly proposed models. For the ICVL dataset \cite{arad2016sparse}, Galliani \etal \cite{galliani2017learned} used a 50\% global split of the available images and randomly sampled a set of $64 \times 64$ image patches for training the Tiramisu network. At test time, they constructed the spectral signatures of a given image by tiling $64 \times 64$ patches with eight pixel overlap to avoid boundary artifacts. For the same dataset, Alvarez-Gila \etal \cite{alvarez2017adversarial} train their network by using a different global split and report their results, making it difficult to validate other approaches due to the lack of uniformly accepted data splits. For AeroCampus, we follow a simple split (Fig.~\ref{fig:aeroRITone}): we use 60 \% of the data as training and the remaining 40 \% as testing. This is done to ensure that there is enough spectral variety present in the dataset with respect to key areas of classes like cars, roads, vegetation and buildings.

\begin{figure*}[t]
\begin{center}
\includegraphics[width= \textwidth, height = 9cm]{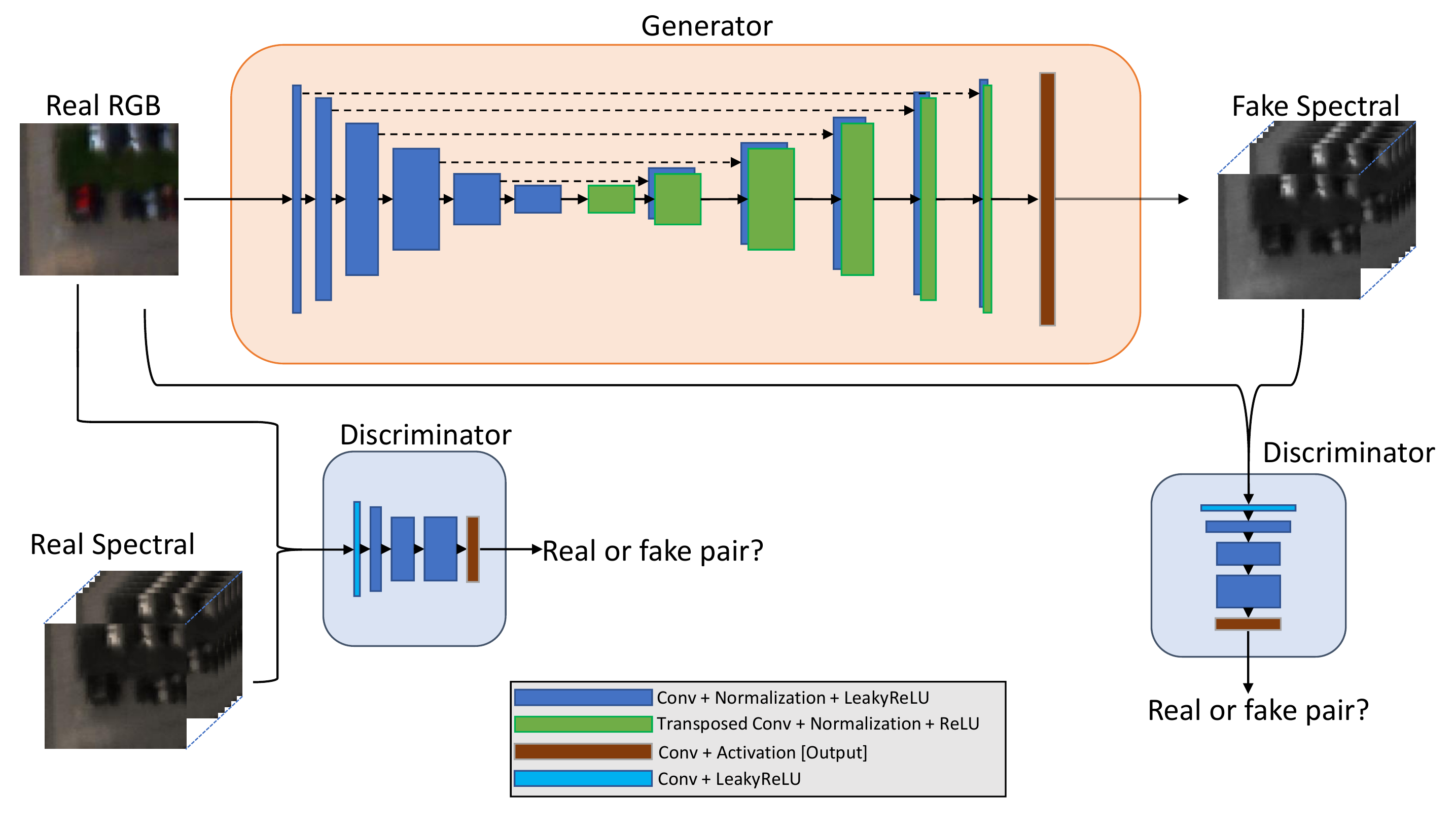}
\end{center}
   \caption{Overall representation of the network. The generator consists of an UNet architecture with 6 up down transitions followed by a 31 band $1 \times 1$ convolution layer and sigmoid activation to predict the output. The discriminator is then used to determine which pair of RGB and Spectral images is real and fake.}
\label{fig:algo}
\end{figure*}

\section{AeroGAN for Aerial SSR}

\textbf{Problem statement.}
As shown in Fig.~\ref{fig:flow}, we define our under-constrained problem as follows: Given a three band (RGB) image, is it possible to learn up-sampling in the spectral domain to regress information for 31 bands between 400 nm - 700 nm? To this end, we experiment with a conventional encoder-decoder network and extend the capacity by modeling the task as a target distribution learning problem.    

\subsection{CNN Framework Analysis}

The network architecture for aerial SSR is constrained by the following requirements: (1) It should be able to process low resolution features very well due to the nature of the data, (2) it should be able to propagate information to all layers of the network so that valuable information is not lost during sampling operations and, (3) it should be able to make the most out of limited data samples. For our model, we use a variant of the UNet \cite{ronneberger2015u} framework since it has been known to operate well on low resolution medical imagery and limited data samples. The network is modified to solve a regression problem by replacing the last softmax layer with a ReLU activation which then gets forwarded to another convolution layer for predicting the band values. The skip connections from encoder to decoder layers ensure conveyance of trivial but useful information whose positioning remains consistent at the output end as well, ensuring all possible information has been utilized to its maximum.


Following popular approaches in spatial super-resolution, we use \textit{Leaky}ReLUs in the encoder side and normal ReLUs in the decoder side to avoid facing vanishing gradients. The last obtained set of filters is then given to a $31$ channel $1 \times 1$ convolution layer \cite{lin2013network} to obtain the final set of 31 bands. The intuition behind using $1 \times 1$ filter here is two fold: it forces the network to learn dimensionality reduction on the 64 channel space and at the same time, gives each of the pixel location its own distinct signature since the filters do not concern themselves with correlation in the spatial feature map space, but rather look at variation in the temporal feature map space. We regress the values for the bands between $0$ and $1$ and found this to be important for achieving a more stable flow in predictions generated by the network. Dropout is applied to all but the last two layers of the CNN to ensure smooth gradient flow through the network while trying to minimize the loss. It is worth mentioning that both, FC-DenseNet (used in \cite{galliani2017learned}) and UNet failed to obtain a good representation of the mapping using conventional loss functions, possibly due to an insufficient number of  training samples. 

\subsection{cGAN Framework Analysis}

While using pixel-wise L1/ MSE loss works for regressing for optimal values of the spectral bands, we further improved the network by turning the problem to a target distribution learning task. Conditional GANs, first proposed in \cite{mirza2014conditional}, have been used widely for generating realistic looking synthetic images \cite{johnson2016perceptual,zhang2016stackgan,ledig2016photo,isola2016image}. To overcome the difficulty of dealing with pixel-wise MSE loss, Johnson \etal \cite{johnson2016perceptual} and Ledig \etal \cite{ledig2016photo} used similar loss functions that were based on the activations of the feature maps in the VGG \cite{simonyan2014very} network layers. There exists no such network in the spectral domain that can help minimize the activations at feature map levels to improve the quality of the generated samples. The functioning of our GAN is inspired by the image to image translation framework of Isola \etal in \cite{isola2016image}. Similar to the their paper where the task is to regress 2/3 channels depending on the problem, we formulate our objective for regressing 31 spectral bands as follows: 
\\
\begin{equation}\label{eq1}\begin{split}
\mathcal{L}_{rgb2si} = \mathbb{E}_{rgb, si \sim p_{data}(rgb, si)} [log \: D(rgb, si)]
\\ 
+ \mathbb{E}_{rgb \sim p_{data}(rgb)} [log \:(1 - D(rgb, G(rgb))] 
\end{split}
\end{equation}

\begin{equation} \label{eq2}
\ G^* = \arg \min_{G} \max_{D} \mathcal{L}_{cGAN} (G,D) + \lambda \mathcal{L}_{other} (G)  
\end{equation}
where the generator (G) tries to minimize the objective $\mathcal{L}_{cGAN} (G,D)$ while the adversarial discriminator (D) tries to maximize it. The other loss in Eqn. \ref{eq1} is an additional term imposed on the generator, which is now tasked with not only fooling the discriminator but also being as close to the ground truth output image as possible. This is accomplished by using $\mathcal{L}_{other}$ as a L1 loss, after having tested with L2 loss and similarity index based losses like SSIM \cite{wang2004image}. L2 loss has been the most popular for pixel-wise reconstruction and though it is effective in low frequency content restoration, it suppresses most of the high frequency detail, which is undesirable given the lack of high frequency content available in the first place. Isola \etal \cite{isola2016image} proposed to trade-off the L2 loss by using L1 loss for correcting low frequency components while using the PatchGAN discriminator to deal with high frequency components by penalizing structural integrity at the patch level. PatchGAN is described in \cite{isola2016image} as the size of the discriminator's receptive field to determine whether that portion of the sample is real or fake. For instance, a $1 \times 1$ receptive field will bias its opinion only on the pixel values individually while a $16 \times 16$ receptive field will determine if the $16 \times 16$ region in the image rendered is real or fake and then average all the scores. This architecture works in our favor since the PatchGAN layers assess spectral data similarity inherently without the need to mention any separate loss function. On the generator side, $\lambda$ is set to $100$ in Eqn. \ref{eq2} with L1 loss to normalize it's contribution in the overall loss function. We found that the best results were obtained (Table \ref{discrimfield}, Fig. \ref{fig:detailanalysis}) by setting the discriminator's receptive field to $70 \times 70$. 

\begin{table}[h]
\centering
{\renewcommand{\arraystretch}{1.2}
\begin{tabular}{lrr}
\hline
\multicolumn{1}{c}{\multirow{2}{*}{\textbf{\begin{tabular}[c]{@{}l@{}}Receptive Field\\ of the Discriminator\end{tabular}}}} & \multicolumn{2}{c}{\textbf{AeroCampus}} \\  
\multicolumn{1}{c}{} & RMSE & PSNR (dB) \\ \midrule
\textbf{$1 \times 1$} & 4.23 & -0.2422 \\
\textbf{$16  \times  16$} & 3.36 & -0.0659 \\
\textbf{$34  \times  34$} & 3.56 & 0.1728 \\
\textbf{$70  \times  70$} & \textbf{2.48} & \textbf{2.0417} \\ \bottomrule
\end{tabular}%
}
\caption{Average root mean square error (RMSE) and peak signal to noise ratio (PSNR) scores for different receptive fields used for the discriminator, evaluated on the test dataset using synthesized RGB patches as inputs to the generator.}
\label{discrimfield}
\end{table}

\section{Experiments and Results}
\label{exp}
\textbf{Data Preparation.} Finding the right alignment between RGB and HSI imagery captured at different altitudes is quite a task when it comes to problems such as SSR. Following the work of other researchers \cite{arad2016sparse,galliani2017learned,dian2017hyperspectral}, we synthesize the RGB images from the hyperspectral data using the standard camera sensitivity functions for the Canon 1D Mark III as collected by Jiang \etal \cite{jiang2013space}. This eliminates the process of establishing accurate spatial correspondence that would have been needed in the original scenario. Camera sensitivity functions give a mapping for the image sensor's relative efficiency of light conversion against the wavelengths. They are used to find correspondences between the radiance in the actual scene and the RGB digital counts generated. In our case, the original hyperspectral scene contains images taken with 372 narrow filters, each separated by about 1 nm. Using ENVI (Exelis Visual Information Solutions, Boulder, Colorado), we first convert this data to 31 bands separated by 10 nm and ranging from 400 nm to 700 nm to form our hyperspectral cube. Using the camera sensitivity function at the corresponding 31 wavelengths, we then synthesize the RGB images. All images are normalized between $0$ to $1$ before being fed into the networks.
\subsection{Settings}

\textbf{Implementation details.} We used PyTorch for all our implementations. All models were initialized with HeUniform \cite{he2015delving} and a dropout of $50 \%$ was applied to avoid overfitting and as a replacement for noise in adversarial networks. For optimization, we used Adam \cite{kingma2014adam} with a learning rate of $2e^-3$, gradually decreasing to $2e^-4$ halfway through the epochs. We found these to be the optimum parameters for all our results. All GANs were trained for 50 epochs to achieve optimal results. All max pooling and up-sampling layers were replaced with strided convolutions and transposed convolutions layers respectively. Inspired by Galliani \etal \cite{galliani2017learned}, we replaced all transposed convolutions with subpixel up-sampling \cite{shi2016real}, but did not achieve significant improvement. Thus transposed convolutions are retained in all our models. 

\textbf{Error metrics.} We use two error metrics for judging the performance of our network: Root Mean Square Error (RMSE) and Peak Signal to Noise Ratio (PSNR). To avoid any discrepancy in the future, it is worth mentioning that the RMSE is computed on a 8-bit range by converting the corresponding values between $[0-255]$ (following approaches in \cite{arad2016sparse,galliani2017learned}) while the PSNR is measured in the $[0-1]$ range.

\subsection{Results}

\begin{figure}[!ht]

\begin{subfigure}[t]{0.23\linewidth}
\centering
\includegraphics[width=\linewidth]{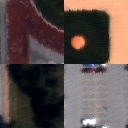}
\end{subfigure}
~
\begin{subfigure}[t]{0.23\linewidth}
\centering
\includegraphics[width=\linewidth]{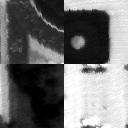}
\end{subfigure}
~
\begin{subfigure}[t]{0.23\linewidth}
\centering
\includegraphics[width=\linewidth]{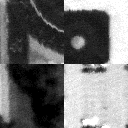}
\end{subfigure}
~
\begin{subfigure}[t]{0.23\linewidth}
\centering
\includegraphics[width=\linewidth]{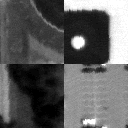}
\end{subfigure} 

\begin{subfigure}[t]{0.23\linewidth}
\centering
\includegraphics[width=\linewidth]{Images/RGB_test.png}
\end{subfigure}
~
\begin{subfigure}[t]{0.23\linewidth}
\centering
\includegraphics[width=\linewidth]{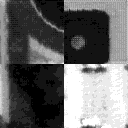}
\end{subfigure}
 ~   
\begin{subfigure}[t]{0.23\linewidth}
\centering
\includegraphics[width=\linewidth]{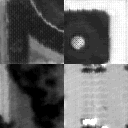}
\end{subfigure}
~ 
\begin{subfigure}[t]{0.23\linewidth}
\centering
\includegraphics[width=\linewidth]{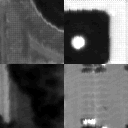}
\end{subfigure}

\caption{Figures showing performance of $70 \times 70$ UNet GAN for the synthesized RGB aerial capture. The first row corresponds to the ground truth while the second row are the model predictions at 420 nm, 550 nm and 620 nm. The networks learns to predict spectral information well, even for cases it has not seen in the training data (running-track and baseball field). }\label{fig:results}
\end{figure}

\begin{figure}[h]
\includegraphics[width=1.0\linewidth]{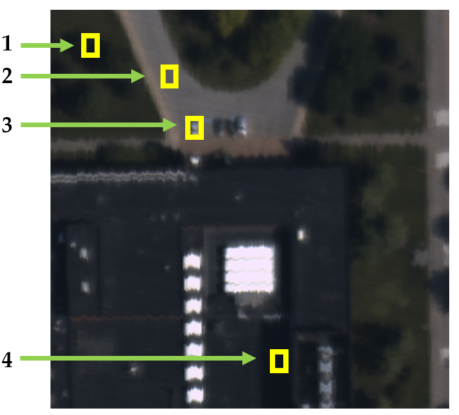}
\caption{Set of points sampled for comparing the spectral distributions in Fig. \ref{fig:detailanalysis}}
\label{fig:sampled}
\end{figure}

\begin{figure}[ht]
\centering
\begin{subfigure}[h]{0.85\linewidth}
\includegraphics[trim={1cm 0 0 0},height=4.7cm, width=1.1\linewidth]{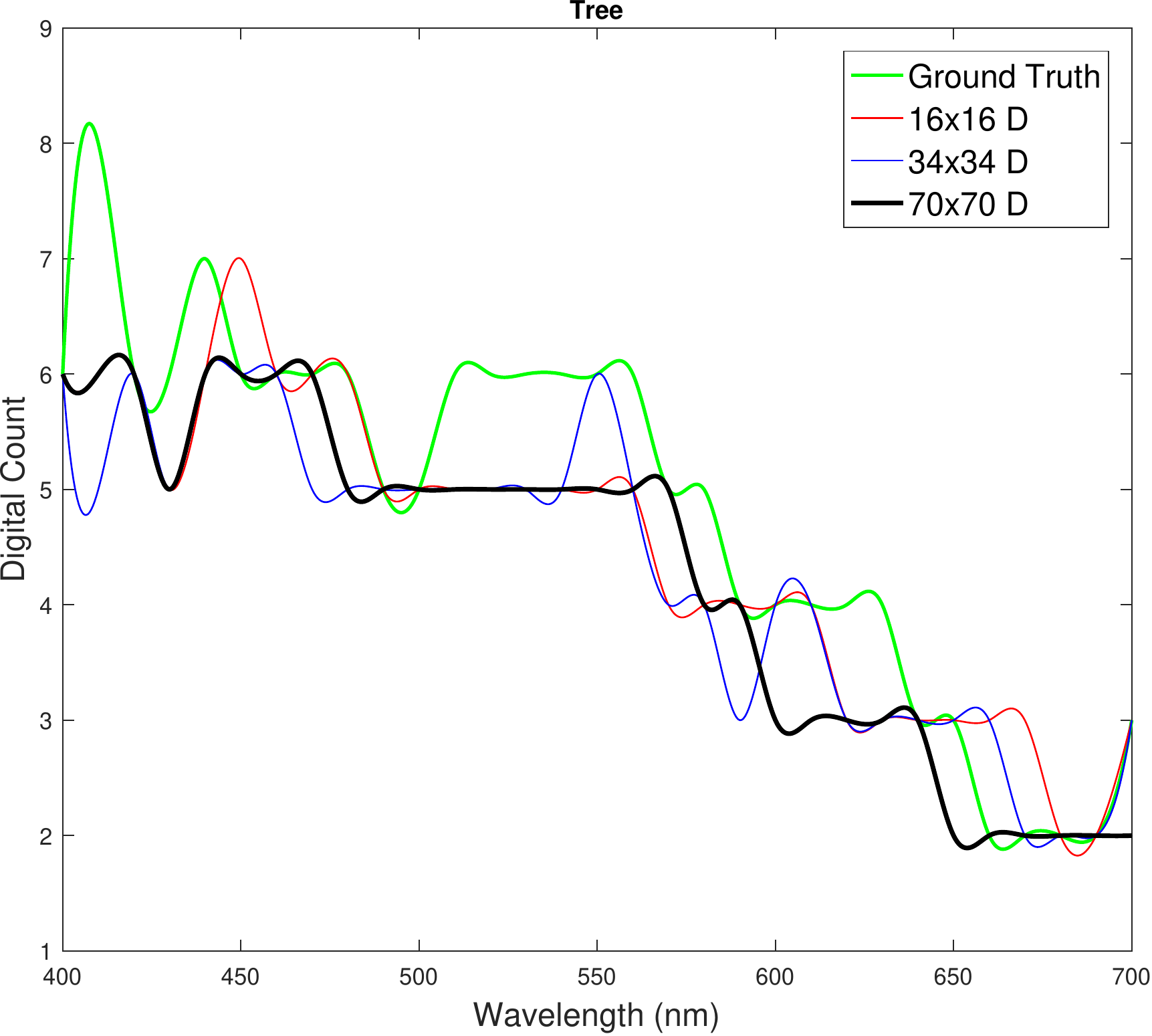}
\end{subfigure}
\begin{subfigure}[h]{0.85\linewidth}
\includegraphics[trim={1cm 0 0 0},height=4.7cm, width=1.1\linewidth]{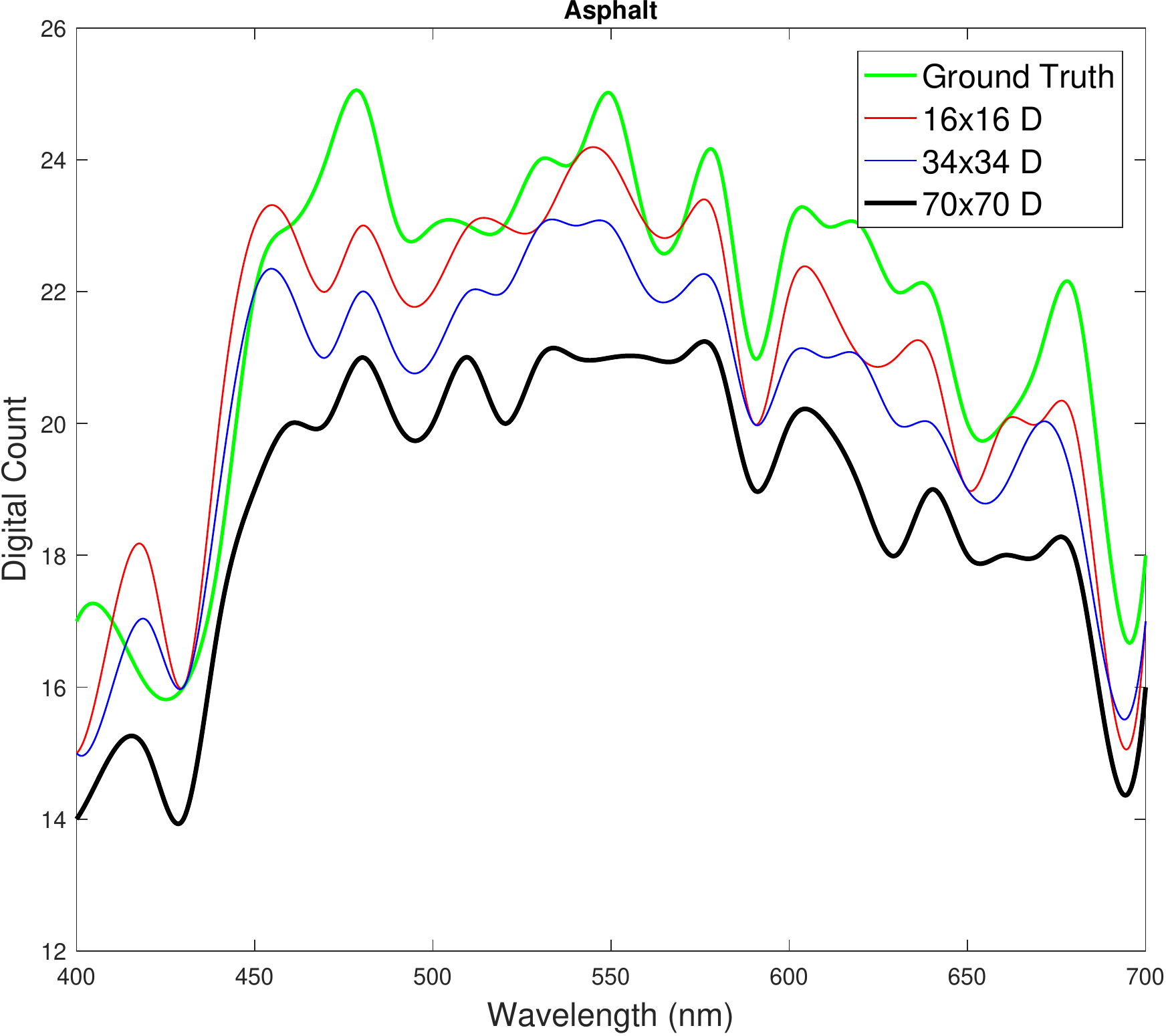}
\end{subfigure}
\begin{subfigure}[h]{0.85\linewidth}
\includegraphics[trim={1cm 0 0 0},height=4.7cm, width=1.1\linewidth]{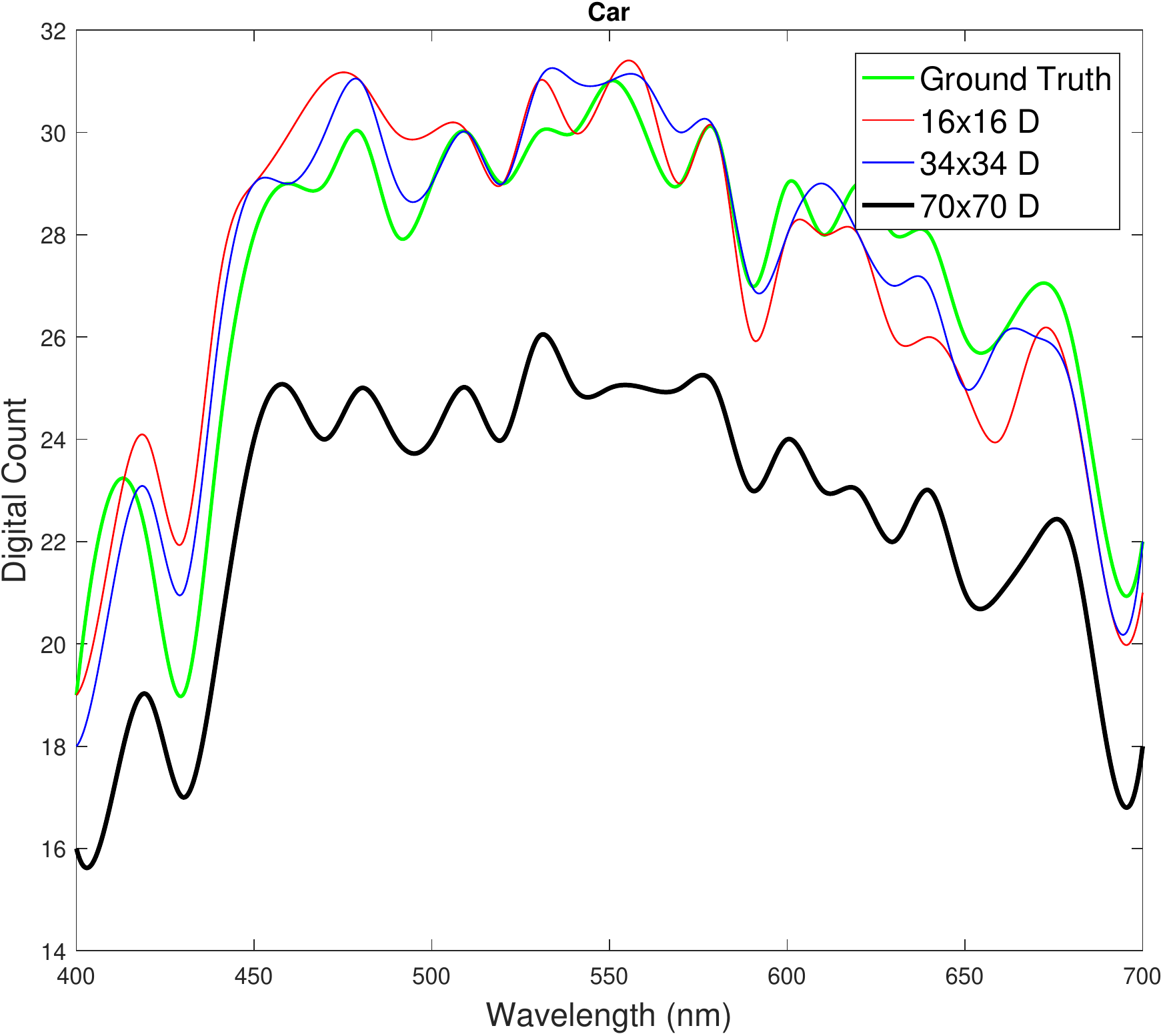}
\end{subfigure}
\begin{subfigure}[h]{0.85\linewidth}
\includegraphics[trim={1cm 0 0 0},height=4.7cm, width=1.1\linewidth]{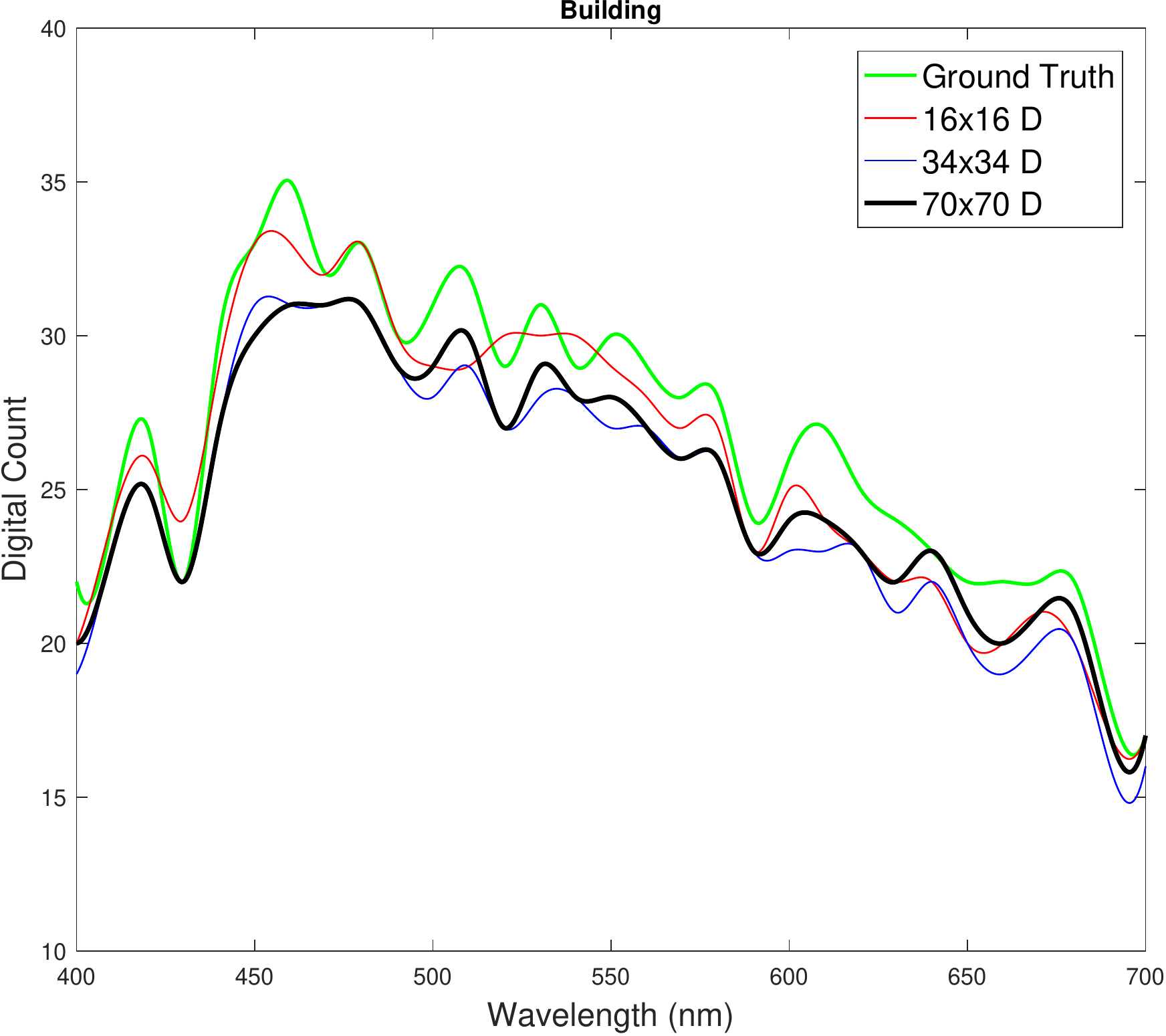}
\end{subfigure}
\caption{An analysis of different spectra sampled from the image (a). The four rests of points correspond to: (1) Tree, (2) Asphalt (road), (3) Car and, (4) Building. The values between the bands have been interpolated by B-spline transform and normalized between $[0-255]$ for analysis.}
\label{fig:detailanalysis}
\end{figure}

Fig. \ref{fig:results} shows a set of different scenarios from the test data that were analyzed. The first row is a set of 4 different scenes from the test dataset at $64 \times 64$ resolution, namely: running track, baseball field, vegetation and parking lot. The scenes are picked such that the former two objects have never been seen by the network and the latter two are some permutation of instances in the training data. The network is able to generate significant band resemblances in all cases, thus proving the viability of our method. Secondly, since the network is fully convolutional, we also test a scenario where it has to infer information in a $256 \times 256$ resolution patch (Fig.~\ref{fig:detailanalysis}). We sample a set of four points as shown Fig. \ref{fig:sampled} and analyze the plots for the three discriminator windows: $16 \times 16$, $34 \times 34$, and $70 \times 70$. 

From Fig. \ref{fig:detailanalysis}, we observe that none of the models predicted the bump observed at 400 - 420 nm range in case of the tree sample. This bump has been caused mostly due to high signal to noise ratio at the sensor end and hence can be treated as noise, which the networks managed to ignore. The inference for car, building and asphalt also looks smooth, and even though the $70 \times 70$ discriminator does not get the  right magnitude levels, the spectra constructed has similar key points for unique object identification, which is close to solving the reconstruction task.

\begin{figure}[t]
\centering
\includegraphics[width=\linewidth]{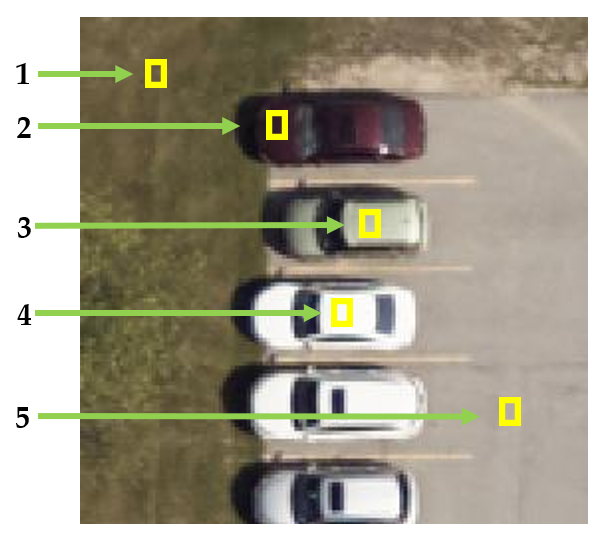}
\caption{A $256 \times 256$ patch sampled from the high resolution RGB image collection towards predicting hyperspectral signatures. A set of 5 points were sampled to assess the performance of the model.}
\label{fig:proof1}
\end{figure}

\textbf{Proof of concept.} The main aim of this study is to figure out if neural networks can learn spectral pattern distributions that could be applied to high resolution RGB images for getting best of both. For validation, we sample a set of patches from the RGB images that were collected and present a proof of concept (Fig.~\ref{fig:proof1}) towards aerial SSR. As observed, the network managed to obtain significant spectral traits: (1) a bump in higher end of the spectrum for the red car and, (2) a peak in green corresponding to the vegetation patch. This shows that it is indeed possible for neural networks to observe information over time and possibly learn a pattern, provided enough samples are present for training. 

\begin{figure}[h]
\centering
\includegraphics[width=\linewidth, height = 7cm]{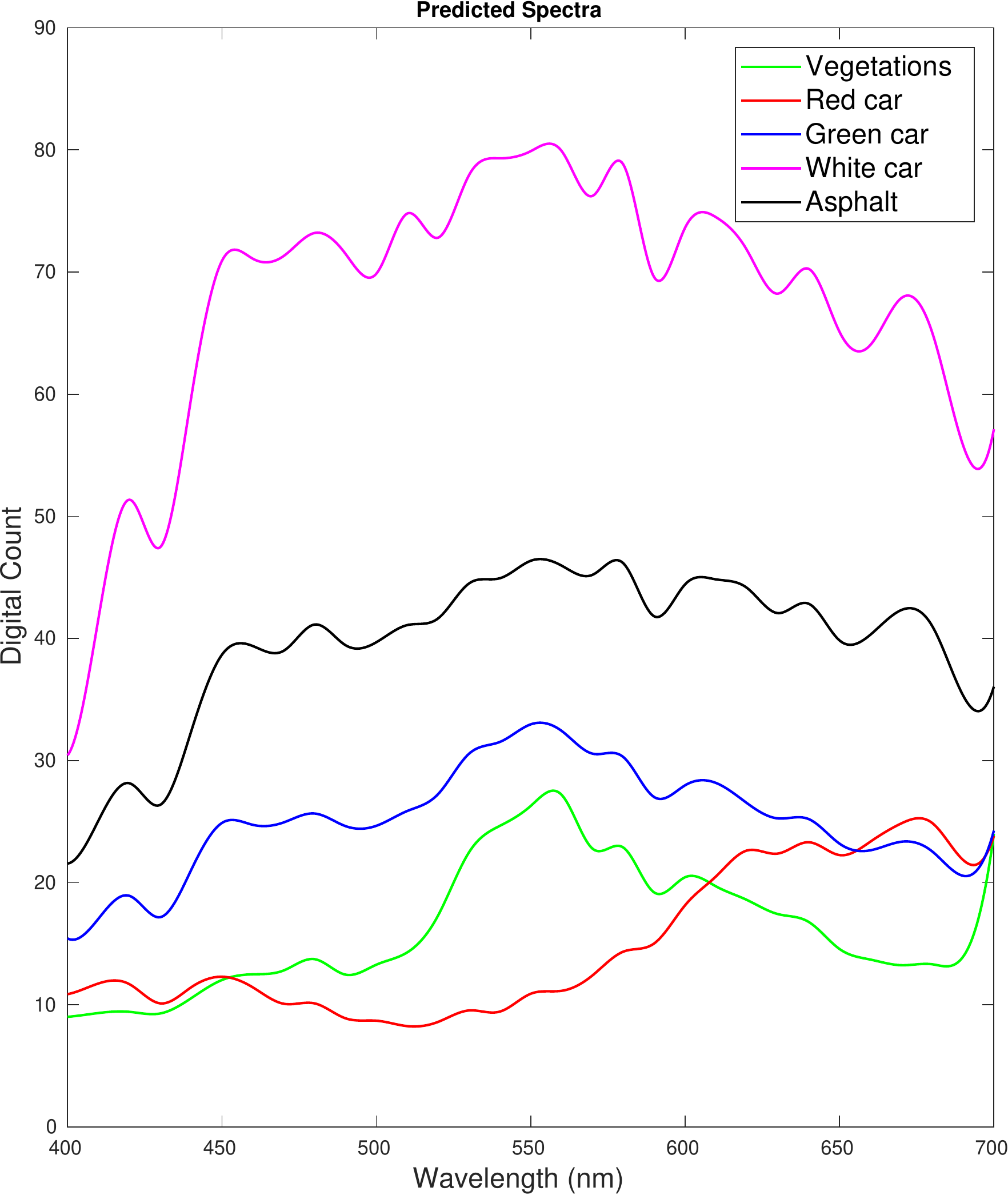}
\caption{Spectral predictions from the $70 \times 70$ GAN for each of the points sampled in Fig. \ref{fig:proof1}. The network has managed to capture traits corresponding to the areas under consideration for most pixels, while confusing between green car (3) and asphalt (5) due to similar RGB combinations. Interestingly, the ground truth values for both the patches are similar with differences in the infrared spectrum, thus in a way proving the network has learnt correctly.}
\label{fig:proofs1}
\end{figure}

\subsection{Discussion}

In this section, we discuss other network architectures that were tried and also the limitations of using SSR with aerial imagery. 

\textbf{Other networks.} Two additional network architectures were tested with to reduce the under-constrained problem space: (1) a 31-channel GAN architecture similar to \cite{suarez2017infrared}, where each band gets its own set of convolution layers before being concatenated for calculating reconstruction loss; and (2) an architecture inspired by \cite{zhang2016stackgan} in which two consecutive GANs learn to first generate an image at a lower resolution ($64 \times 64$) and then upscale to a higher resolution ($256 \times 256$). In our case, we used two different GANs to first spectrally up-sample to 11 bands and then predict the remaining 20. However, we found both these networks to be more unstable than the simpler one. We believe the cause for this to possibly be the fact that it is more easier to learn an entire spectral distribution range than learning it split by split since there can be overlaps between objects of different categories in particular spectral ranges. We are continuing to develop these models. 

\begin{figure}
\centering
\includegraphics[width=\linewidth]{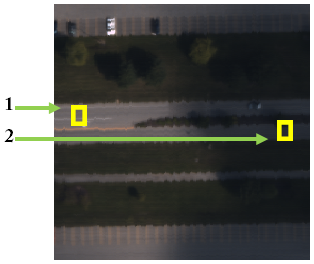}
\caption{A $256 \times 256$ patch sampled from another flight line during occlusion by clouds. Two sets of road patches are sampled from this image: one under sunlight and the other under shadows.}
\label{fig:shadowannot}
\end{figure}

\begin{figure}
\centering
\includegraphics[trim={0.5cm 0 0 0}, width=\linewidth, height = 9cm]{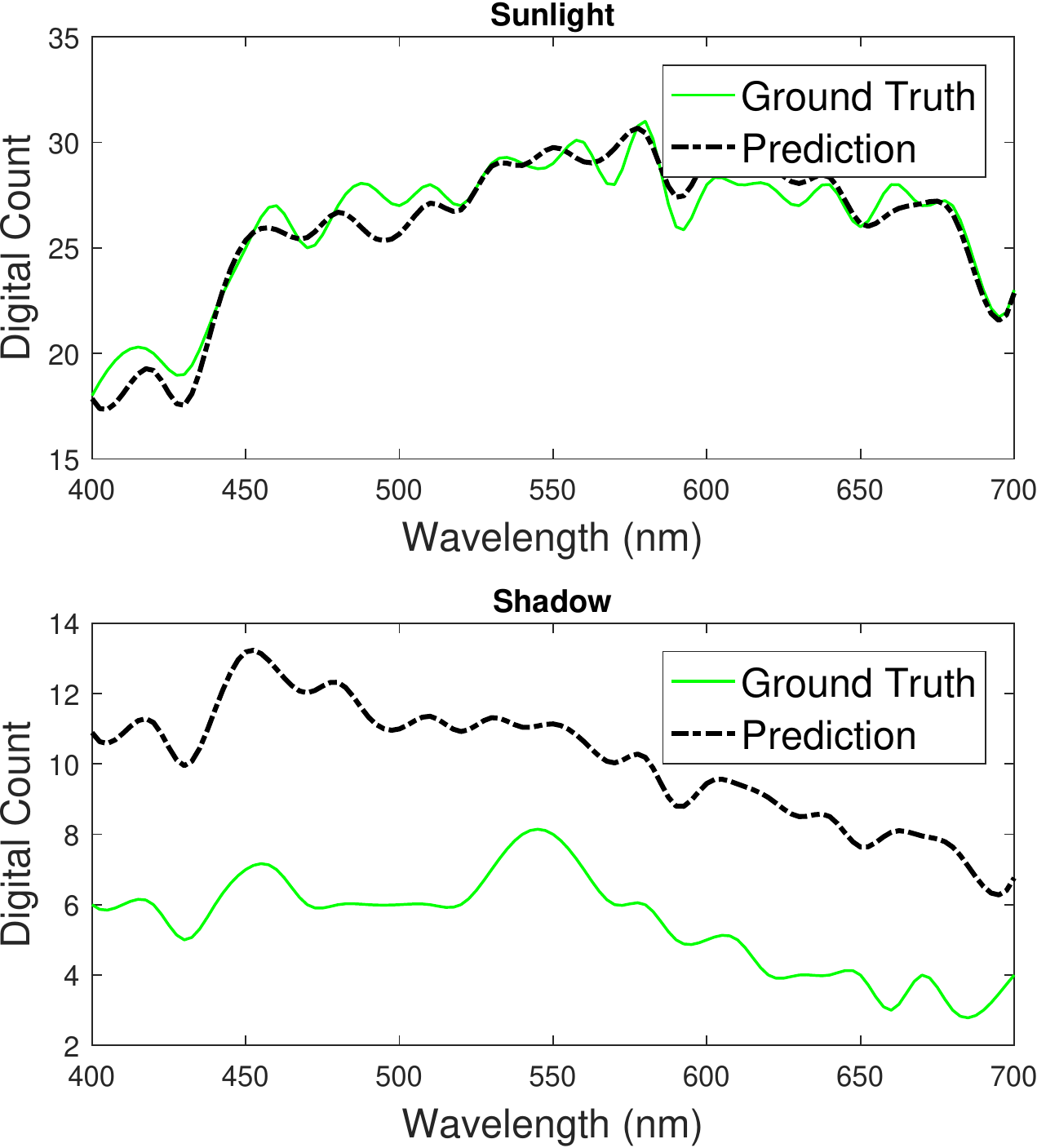}
\caption{Plot showing ground truth and predicted spectral bands for patches sampled in Fig. \ref{fig:shadowannot}. As seen, the spectral prediction for the sunlight patch (1) is pretty accurate, while the network struggles to obtain the right level of magnitude for shadow patch (2).}
\label{fig:shadowsoutput}
\end{figure}

\textbf{Areas of development.} SSR has its own set of limitations that cannot be resolved irrespective of the methods used. For example, one of the main motivations for this paper is to determine if an applied learning can be used instead of expensive hyperspectral cameras to predict light signatures in the hyperspectral space. While it is possible to model spectral signatures between $400$nm - $700$nm, it is next to impossible to model infrared and beyond signatures since they are not a function of just the RGB values. Here, we present two ``solvable'' limitations: Water and Shadows. Water does not have its own hyperspectral signature and instead takes over the signature of the sediments present in it - the signatures for clear water and turbid water would be distinctly apart. Detecting shadows has been known to be a problem in spectral imaging~\cite{huang2015detecting} since they also do not exhibit an unique spectral signature. The question posed here is simple - given a vast amount of data, is it possible to have a network learn how water and shadows work and affect the spectral signatures of objects under consideration? To this end, we sample a $256 \times 256$ patch from another flight line (Fig.~\ref{fig:shadowannot}) that contains asphalt (road) under two different circumstances: sunlight and shadows. The corresponding spectral prediction is shown in Fig.~\ref{fig:shadowsoutput} where we observe that the network managed to have a similar spectral signature to the sunlight patch with a decrease in magnitude. This could be of importance in tasks where knowing the presence of shadows is required. 

\section{Conclusion}
\label{conc}

In this paper, we trained a conditional adversarial network to determine the 31 band visible spectra of a aerial color image. Our network is based on the Image-to-Image Translation framework which we extend to predict 31 band values. We show that the network learns to extract features for determining an object's spectra despite high noise interference in the spectral bands. Experimental results show a RMSE of 2.48, which shows that the network is successfully recovering the spectral signatures of a color image. Furthermore, we introduce two modeling complexities: water and shadows and release the AeroCampus dataset for other researchers to use.

\section{Acknowledgments}
\label{ac}
This work was supported by the Dynamic Data Driven Applications Systems Program, Air Force Office of Scientific Research, under Grant FA9550-11-1-0348. We thank the NVIDIA Corporation for the generous donation of the Titan X Pascal that was used in this research.

{\small
\bibliographystyle{ieee}
\bibliography{egbib}
}

\end{document}